\pdfoutput=1

\documentclass[11pt]{article}

\usepackage[final]{acl}

\usepackage{times}
\usepackage{latexsym}

\usepackage[T1]{fontenc}

\usepackage[utf8]{inputenc}

\usepackage{microtype}

\usepackage{inconsolata}

\usepackage{graphicx}
\usepackage{amsmath}
\usepackage{multirow}

\newtheorem{assumption}{Assumption}

\title{Challenging Assumptions in Learning Generic Text Style Embeddings}

\author{Phil Ostheimer, Marius Kloft, Sophie Fellenz \\
RPTU Kaiserslautern-Landau \\
surname@cs.uni-kl.de\\}

\begin{document}
\maketitle
\begin{abstract}
Recent advancements in language representation learning primarily emphasize language modeling for deriving meaningful representations, often neglecting style-specific considerations. This study addresses this gap by creating generic, sentence-level style embeddings crucial for style-centric tasks. Our approach is grounded on the premise that low-level text style changes can compose any high-level style. We hypothesize that applying this concept to representation learning enables the development of versatile text style embeddings. By fine-tuning a general-purpose text encoder using contrastive learning and standard cross-entropy loss, we aim to capture these low-level style shifts, anticipating that they offer insights applicable to high-level text styles. The outcomes prompt us to reconsider the underlying assumptions as the results do not always show that the learned style representations capture high-level text styles.
\end{abstract}

\section{Introduction}
Learning meaningful representations of text has received much attention recently as models pre-trained on large corpora have become the standard for extracting representations that capture prior knowledge (i.a. \citealp{Devlin:2019, Radford:2018}). However, most pre-trained models focus on general aspects, as they are trained with causal or masked language modeling objectives. Therefore, they are suboptimal for NLP tasks focusing on a specific aspect, such as style. There are many notable NLP tasks, such as identifying chatbot-written text based on its style (e.g., \citet{Soto:2024}) or style transfer where models rely on dedicated style representations (e.g., \citet{John:2019}). 

Existing work on style transfer focuses on high-level stylistic aspects such as sentiment \cite{Shen:2017} or formality \cite{Rao:2018}, taking style definitions based on the dataset's structure. \citet{Lyu:2021} view those high-level text style changes as compositions of more fine-grained, low-level style changes. Generic style representations incorporating low-level aspects such as lexical, syntactic, semantic, and thematic stylistic traits \cite{McDonald:1985, DiMarco:1993} but also high-level, composed stylistic traits might significantly improve style-focused tasks.

Based on the hypothesis that low-level style changes compose high-level style changes, this work explores learning generic, sentence-level style representations. We take a pre-trained encoder model producing general-purpose representations and fine-tune it to distinguish between low-level stylistic changes using contrastive learning and cross-entropy loss. We hypothesize that the resulting text encoder generalizes to high-level styles, applying the view by \citet{Lyu:2021} on how low-level stylistic changes compose high-level changes to representation learning for styles.

We train our method using contrastive learning and cross-entropy loss on the StylePTB dataset \cite{Lyu:2021} comprising low-level, fine-grained style changes to obtain high-level, generic style embeddings. We evaluate our method by training a simple classifier on the representations of the learned style. The results show an ambiguous picture of the resulting style embeddings, challenging the underlying assumptions.

\section{Related Work}
\paragraph{Explicitly Learning Style Representations}
Only a few works learn style representations explicitly. StyleDistance \citep{Patel:2024} uses a contrastive triplet loss and synthetic parallel data created by Large Language Models to learn generic style representations, showing strong performance in multiple benchmarks.

\paragraph{Style Representations as a Byproduct} 
Text style transfer is conducted by prominent models by disentangling content and style to learn separate task-specific representations to control them independently \cite{Fu:2018, Hu:2017, Kim:2020, John:2019, Cheng:2020}. Other text style transfer models learn content representations and multiple decoders (one for each style) \cite{Shen:2017, Fu:2018}. Another group of models uses a structured style code to enforce a particular style in the decoder, either given as a structured code \cite{Hu:2017, Lample:2019} or learned \cite{Fu:2018, Kim:2020}.

For text style classification, TextCNN \cite{Kim:2014} is the most widely used method \cite{Ostheimer:2023}. BERT \cite{Devlin:2019} and RoBERTa \cite{Liu:2019}, fine-tuned for style classification, are strong baselines.

In contrast, our proposed method learns style representations at the sentence level from low-level, linguistically motivated style changes. This is much more fine-grained and allows applications to unseen styles.

\paragraph{Contrastive Learning for Text Representations}
To learn meaningful content representations on the sentence level from unlabeled text corpora, the QT model \cite{Logeswaran:2018} was introduced. The QT model relies on the distributional hypothesis to get meaningful content representations. It uses a contrastive objective to map nearby (context) sentences to similar and distant (non-context) sentences to far-apart representations.

For fine-tuning sentence representations, notable approaches are SimCSE \citep{Gao:2022} and Mirror-BERT \cite{Liu:2021c}, incorporating minimal data augmentation with dropout. In contrast, SBERT \citep{Reimers:2019} uses siamese and triplet network structures to generate meaningful sentence representations for calculating semantic sentence similarities using standard measures like cosine distance. \citet{Kim:2021} improve the quality of the sentence representations by contrasting the representations of different layers of BERT \cite{Devlin:2019}.

However, these approaches focus on improving the sentence-level representations for general language understanding or semantics. We, in contrast, focus on the style of the sentences.

\section{Method}
\begin{figure}[t]
\begin{center}
\includegraphics[width=\columnwidth]{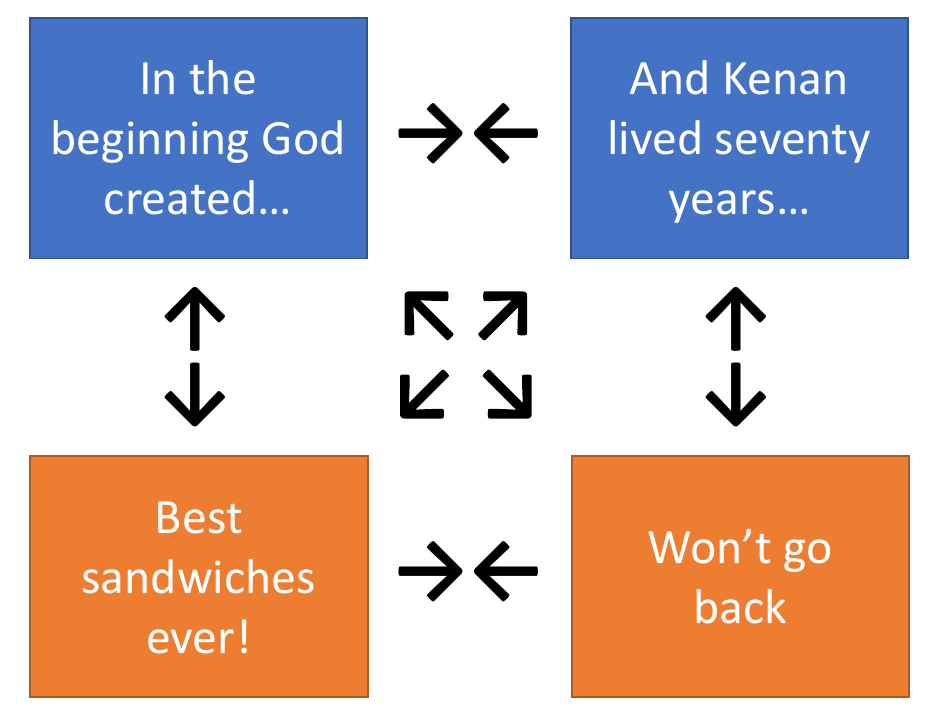} 
\caption{Our training objective pushes sentence representations of the same style close together. In this example, reviews (in orange) are pushed close together, and sentences of one Bible version (in blue) are pushed close together, while the representations of different styles (Bible vs reviews) are pushed to be far apart.}
\label{fig:model_diagram}
\end{center}
\end{figure}

This section describes the underlying assumptions of the proposed method to compute (sentence-level) style representations of texts and the method itself. 

\begin{assumption} 
\label{ass:low_high}
Low-level stylistic changes can be composed to form high-level styles in line with previous work \citep{Lyu:2021}.
\end{assumption}

\begin{assumption} 
\label{ass:contr}
Learning style embeddings can be achieved by contrasting different styles \citep{Patel:2024}.
\end{assumption}

We combine the aforementioned assumptions to come up with the following method.

\paragraph{Method} The core idea---illustrated in Figure \ref{fig:model_diagram}---is to embed the data into a space where texts of the same style resemble each other, while texts of different styles are easily distinguishable. Formally, we want to learn a neural encoder $f$ mapping any two input sentences $x$ and $\tilde{x}$ onto vector representations $f(x)$ and $f(\tilde{x})$ such that $f(x)\approx f(\tilde{x})$ if and only if $x$ and $\tilde{x}$ are of the same style: $\mbox{style}(x)=\mbox{style}(\tilde{x})=s$. To achieve this, we first form pairs of sentences, some equal and others of a different style. Then, we use a contrastive objective to push the vector representations of sentences of the same style closer together and ones of different styles far apart. 

For a given sentence $x$ with $\mbox{style}(x)=s$, our set of candidate sentences $X_{\text{cand}}$ contains sentences of the same style and different style(s):  $X_{\text{cand}} = X_{s} \cup \overline{X_{s}}$. We employ a simple architecture to compute the similarity between the outputs $f(x)$ and $f(\tilde{x})$, namely the inner product $f(x)^\top f(\tilde{x})$. Other functions to consider are, e.g., the cosine similarity or kernel functions. We compute the similarity using a simple architecture to avoid learning a rich similarity measure compensating for the encoder producing poor style representations.

Therefore, we have the following  formulation for the probability of each candidate sentence $\tilde{x} \in X_{\text{cand}}$ to have the same style as $x$:
\begin{equation}
    p(\tilde{x}|x,X_{\text{cand}})=\frac{\exp[ f(x)^\top f(\tilde{x})]}{\sum_{\tilde{x}' \in X_{\text{cand}} \exp[f(x)^\top f(\tilde{x}')]}}
\end{equation}

Our training objective is to maximize the probability of identifying all sentences $\tilde{x} \in X_{\text{cand}}$ where $\mbox{style}(x)=\mbox{style}(\tilde{x})=s$ for each sentence $x$ in the training data $D$:
\begin{equation}
    \sum_{x \in D} \sum_{\tilde{x} \in X_{s}} \log p(\tilde{x}|x,X_{\text{cand}})
\end{equation}

We also experiment with a cross-entropy loss and its combination with a contrastive loss.

\section{Experiments}
In this section, we describe how we evaluated the effectiveness in learning style representations for multiple styles.
\begin{table*}[ht]
\begin{center}
\begin{tabular}{c|l|c|c|cc|cc|cc}
\multicolumn{4}{c}{} & \multicolumn{2}{|c|}{Random Sampler} & \multicolumn{2}{c|}{Pairwise Sampler} & \multicolumn{2}{c}{Corpus Sampler}\\
& Dataset & PT & CEL & CEL+CL & CL & CEL+CL & CL & CEL+CL & CL\\
\hline
\multirow{5}{*}{\rotatebox{90}{BERT}}& Yelp & \textbf{94.3} & 92.9 & 90.5 & 90.1 & 72.9 & 87.0 & 71.4 & 88.1\\
& Amazon & \textbf{77.4} & 74.9 &  72.1 & 74.5  & 59.6 & 68.3 & 58.1 & 69.3 \\
& GYAFC & \textbf{88.4} & 88.0 & 82.1 & 82.1  & 71.5 & 86.2 & 74.3 & 78.3\\
& Shakespeare & 85.7 & \textbf{86.4} & 83.4 & 83.5  & 75.1 & 83.7 & 73.7 & 81.7 \\
& Bible & 50.9 & \textbf{52.8}& 42.9 & 46.7 & 31.1 & 52.4  & 33.9 & 41.2\\
\hline
\multirow{5}{*}{\rotatebox{90}{RoBERTa}} & Yelp & 86.2 & \textbf{87.5} & 85.3 & 74.2 & 55.1 & 50.8  & 56.0 & 71.9\\
& Amazon & \textbf{75.2} & 67.1 & 64.4 & 60.3 & 54.1 & 65.4 & 55.9 & 58.1 \\
& GYAFC & 79.4 & \textbf{85.5} & 80.2 & 76.3 & 55.5 & 59.0 & 60.4 & 66.9\\
& Shakespeare &  80.1 & \textbf{85.1} & 81.7 & 76.1 & 64.8 & 74.0 & 67.4 & 77.6 \\
& Bible & 55.9 & \textbf{60.8} & 53.8 & 44.4  & 18.9 & 24.2 & 19.1 & 38.5\\
\end{tabular}
\caption{Shown is the style classification accuracy of a logistic regression fitted to the two text encoders $f$ BERT and RoBERTa as a pre-trained (PT) encoder, fine-tuned on StylePTB and applied to the mentioned datasets using a Cross-Entropy Loss (CEL), Contrastive Loss (CL), or both   (CEL+CL) with the three mentioned sampling strategies.}
\label{tab:bert_roberta_cl}
\end{center}
\end{table*}

\subsection{Experimental Setup}
\paragraph{Fine-Tuning Data}
We use the StylePTB \cite{Lyu:2021} dataset containing 21 individual and 32 compositional fine-grained style changes to train our generic text style embeddings.

\paragraph{Evaluation Data}
We evaluate the learned style embeddings on datasets that take a data-driven approach, containing high-level stylistic changes, to define a text style. These are used in many recent works on text style transfer. We experiment with the Bible corpus \cite{Carlson:2018} with eight different styles of the Bible (249K sentences), Grammarly's Yahoo Answers Formality
Corpus (GYAFC) \cite{Rao:2018} (113K sentences) in two styles, a collection of Shakespeare plays in Shakespearean and modern English \cite{Xu:2012} (42K sentences), Amazon (558K) and Yelp (448K) sentiment datasets\footnote{https://github.com/lijuncen/Sentiment-and-Style-Transfer} (two styles each). We follow prior work by using the existing train-dev-test splits. 

\paragraph{Training}
We experiment with both $\textrm{RoBERTa}_{\textrm{Large}}$ \cite{Liu:2019} and $\textrm{BERT}_{\textrm{Large}}$ \cite{Devlin:2019} as pre-trained encoders for $f$. The sentence representation is the activation from the last hidden layer for the ``CLS'' token. We add a linear transformation and a $l_2$ normalization before applying the objective function. Training stops after a maximum of 10 epochs, and the best model is used based on the loss on the validation dataset. Hyperparameters were chosen using the loss on the validation dataset. We used a batch size of 16, a learning rate of 1e-5 with a linear warmup for the first 10\% steps, followed by a linear cooldown for the remaining steps. We used an Adam optimizer and a dropout rate of 0.1. A logistic classifier is trained on the training data and evaluated on the test data using the representations obtained by the encoder $f$. 

\paragraph{Batch Construction}
\label{sec:batch}
We initially experimented with contrasting more than two styles per batch. However, this resulted in no meaningful representations. Therefore, we resorted to two styles per batch. We use a random sampler with replacements to randomly select sentences of each style. We also use two random samplers without replacement: one where we assure for each style pair that only unseen sentence pairs are contrasted and one where we only assure on the corpus level that per epoch, the sentences are only contrasted once. Half of the batch consists of style $s$ while the other half contains style $\overline{s}$. 

\paragraph{Obtaining Generic Style Embeddings}
To get generic style embeddings, we fine-tune pre-trained encoders $f$ on StylePTB \cite{Lyu:2021} and apply the resulting encoders to the previously mentioned high-level style datasets. The hypothesis is the following: Since StylePTB contains fine-grained (and compositional) style changes, it should also generalize to high-level and unseen styles.

\subsection{Results}
In Table \ref{tab:bert_roberta_cl}, we summarize our style classification results. Using pre-trained BERT and RoBERTa encoders as $f$, we apply a logistic regression for classification without fine-tuning, serving as our baseline. Baseline accuracies for datasets like Yelp, Amazon, GYAFC, and Shakespeare (two styles each) are already nearly 80\% or higher. The Bible dataset, with eight styles, exhibits lower accuracy, as expected. Generally, BERT outperforms RoBERTa in style classification. Our approach considers three sampling strategies (Section \ref{sec:batch}), crucial for contrastive loss performance.

Fine-tuning BERT and RoBERTa with the cross-entropy loss yields slight accuracy improvements for the Shakespeare and Bible corpora. Moreover, fine-tuning RoBERTa further enhances accuracies on Yelp and GYAFC. However, incorporating the contrastive loss reduces accuracy compared to cross-entropy fine-tuning. Solely fine-tuning with a contrastive loss also leads to less accuracy.

\subsection{Discussion}
While fine-tuning an encoder $f$ on the StylePTB dataset using cross-entropy loss slightly improves some settings' accuracy, contrastive learning does not. These findings of our study question the representational capacity of the learned style embeddings, especially for contrastive learning.
\paragraph{Contrastive Objective Is Too Aggressive} Contrary to our expectations, applying the contrastive loss does not improve the accuracy compared to the cross-entropy loss across various settings we explored. One possible reason is that the contrastive loss might push dissimilar styles too far apart. Contrary to previous work \citep{Patel:2024}, we do not use synthetic parallel data but contrast non-parallel data. This might hamper the model's ability to learn the relationship between different styles and style levels.
\paragraph{RoBERTa's ``CLS'' Tokens Need Fine-Tuning} The improvements with the cross-entropy loss can be attributed to the fact that the ``CLS'' token is not pre-trained using the next sentence prediction task and, therefore, any fine-tuning might improve the ``CLS'' token representations. 
\paragraph{Differentiated Picture for Cross-Entropy Loss} One possible reason for the differentiated results can be seen in the examples from StylePTB in Table \ref{tab:styleptb_examples}. While some low-level changes, like info addition, relate directly to a formality change, others, such as tense changes, do not align with the investigated styles. Mixing low-level styles may confuse the encoding mechanisms that distinguish higher-level styles.

\begin{table}[ht]
\begin{center}
\begin{tabular}{p{1.5cm}|p{2.4cm}|p{2.4cm}}
Aspect & Original & Transferred\\
\hline
Info addition & Morgan Freeman did the new one & Morgan Freeman did  \emph{perform} the new one.\\
\hline
To future tense & It is also planning another night of original series. & It will be also planning another night of original series. \\
\end{tabular}
\caption{Examples from the StylePTB dataset}
\label{tab:styleptb_examples}
\end{center}
\end{table}

\section{Conclusion}

Learning generic, high-level text style representations from low-level, linguistically motivated changes and generalizing to high-level styles according to Assumption \ref{ass:low_high} by \citet{Lyu:2021} using contrastive learning according to Assumption \ref{ass:contr} by \citet{Patel:2024} presents challenges. Although the approach does not yield the expected results using contrastive learning, cross-entropy loss shows improvements for some settings. However, our approach does not yield the expected results compared to previous work \citep{Patel:2024} using contrastive learning to learn generic style representations.

\section*{Limitations}
One limitation of this study is the reliance on the StylePTB dataset, which, to our knowledge, is the only available dataset containing low-level and composed stylistic changes. The dataset is restricted to English, limiting the generalizability of our findings to other languages. As style may manifest differently across languages, a more diverse, multilingual dataset would allow for broader application and a more comprehensive evaluation of the proposed method.

Additionally, our study focuses on contrastive learning to capture text style representations. This choice was made because it intuitively aligns with the underlying assumptions of the task, but it may not be the optimal approach for all settings.

\section*{Acknowledgements}
Part of this work was conducted within the DFG Research Unit FOR 5359 on Deep Learning on Sparse Chemical Process Data (BU 4042/2-1, KL 2698/6-1, and KL 2698/7-1) and the DFG TRR 375 (no. 511263698). SF and MK acknowledge support by the Carl-Zeiss Foundation (AI-Care, Process Engineering 4.0), the BMBF award 01|S2407A, and the DFG awards FE 2282/6-1, FE 2282/1-2, KL 2698/5-2, and KL 2698/11-1.
\bibliography{custom}

\end{document}